\documentclass[11pt]{article}

\usepackage[preprint]{acl}

\usepackage{times}
\usepackage{latexsym}

\usepackage[T1]{fontenc}

\usepackage[utf8]{inputenc}

\usepackage{microtype}

\usepackage{inconsolata}

\usepackage{graphicx}
\usepackage{booktabs}
\usepackage{amsmath}
\usepackage{amssymb}
\usepackage{nicefrac}
\usepackage{multirow}
\usepackage{pgfplots}
\usepgfplotslibrary{groupplots}
\usepackage{algorithm}
\usepackage{algorithmic}

\title{LatentRevise: Learning from Zero-Hit Reasoning}

\author{
  \makebox[\textwidth][c]{%
      \large
      \textbf{Yiqiu Guo}$^{1*}$ \hspace{8pt}
      \textbf{Xueting Han}$^{2\dagger}$ \hspace{8pt}
      \textbf{Qi Jia}$^3$ \hspace{8pt}
      \textbf{Guangtao Zhai}$^{3,4}$ \hspace{8pt}
      \textbf{Jing Bai}$^2$
  } \\[0.4em]
  \makebox[\textwidth][c]{%
      \normalsize
      $^1$Fudan University\quad
      $^2$Microsoft Research Asia
  } \\[0.2em]
  \makebox[\textwidth][c]{%
      \normalsize
      $^3$Shanghai AI Laboratory\quad
      $^4$Shanghai Jiao Tong University
  } \\[0.2em]
  }

\begin{document}
\maketitle
\renewcommand{\thefootnote}{\fnsymbol{footnote}}
  \footnotetext[1]{\footnotesize\raggedright Work was done during Yiqiu Guo's
    \href{mailto:yqguo22@m.fudan.edu.cn}{\texttt{<yqguo22@m.fudan.edu.cn>}} internship at MSRA.}
  \footnotetext[2]{\footnotesize\raggedright Correspondence to: \href{mailto:chrihan@microsoft.com}{\texttt{<chrihan@microsoft.com>}}.}, 
\begin{abstract}
Reinforcement learning with verifiable rewards (RLVR) is bottlenecked by hard prompts on which correct trajectories have low probability, so sampling misses them within a practical budget and leaves the policy update with little useful signal. We frame such zero-hit prompts as RLVR's sampling frontier, where new reasoning behavior is most valuable yet least likely to be sampled. Importantly, failed rollouts can be informative: they expose where the model's reasoning went wrong.
We introduce LatentRevise, a first-order latent revision method that recovers training signal for this zero-hit regime. Given a failed rollout and the gold answer as an anchor, LatentRevise optimizes the input embeddings of its reasoning prefix under two complementary gradients, moving the prefix away from the failed continuation and toward the gold answer.
The optimization is constrained to the convex hull of the model's vocabulary embeddings, so each update moves the latent toward a real token embedding rather than an arbitrary feature direction.
We find that continuations from the revised prefix lengthen, exhibit self-reflection, and reach correct answers missed by the original rollouts. Used as training data, these trajectories improve SFT and RLVR on math benchmarks over standard baselines.
\end{abstract}

\section{Introduction}
\label{sec:intro}
Reinforcement learning with verifiable rewards (RLVR) has become a central recipe for improving reasoning capabilities in language models~\citep{guo_deepseek-r1_2025}. Group-relative methods such as GRPO~\citep{shao_deepseekmath_2024} turn binary feedback into a learning signal across sampled rollouts, increasing the likelihood of verified trajectories and decreasing that of rejected ones. Yet this mechanism is inherently tied to what the current policy can sample under rollout budget. Recent analyses suggest that RLVR often behaves as support-constrained optimization: it sharpens probability mass around high-reward trajectories already reachable from the base policy, and can narrow empirical support under larger sampling budgets~\citep{wu_invisible_2025,yue_does_2025}.

\begin{figure}[t!]
  \centering
  \includegraphics[width=1.0\linewidth]{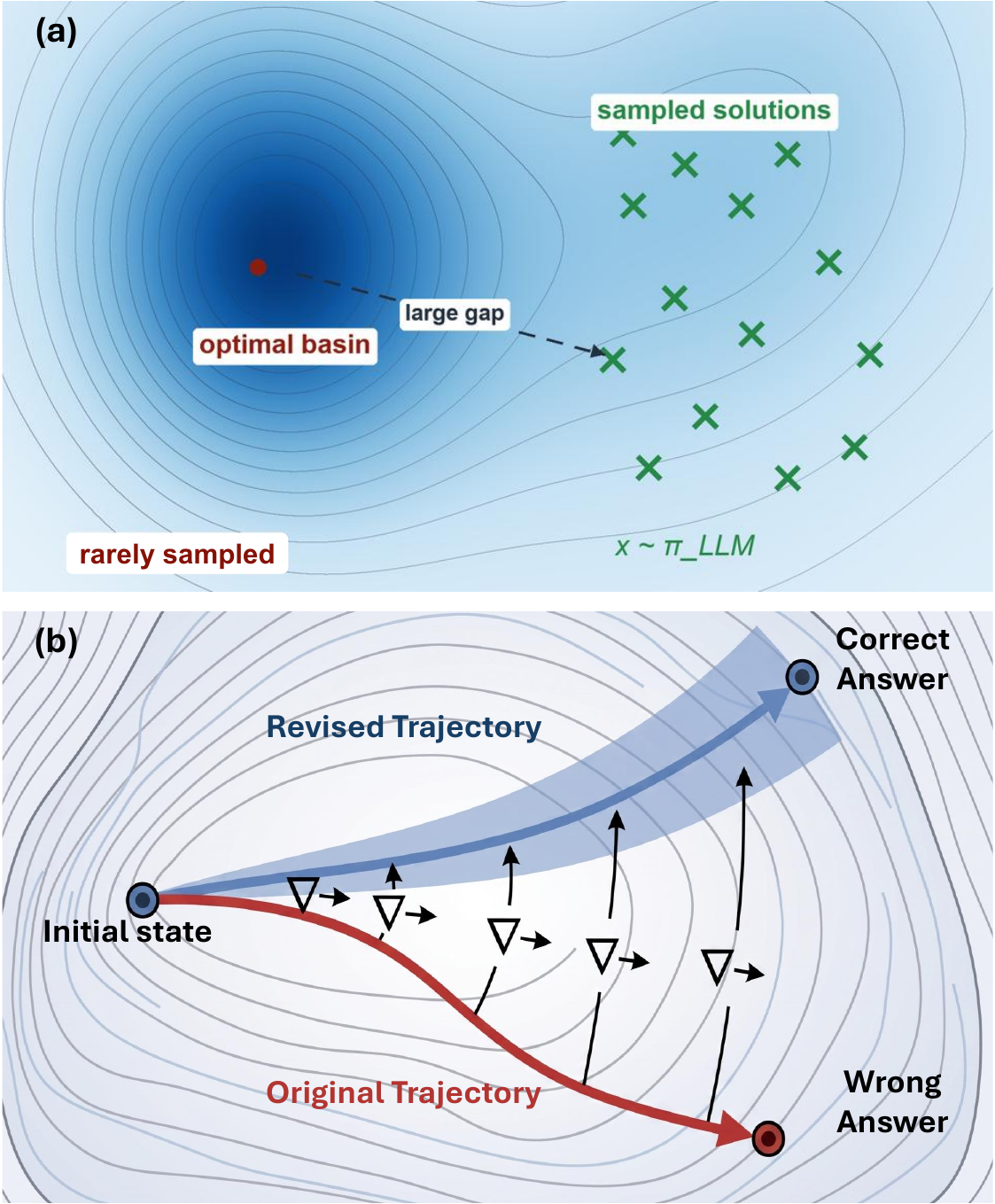}
  \caption{From zero-hit prompts to revised trajectories. (a) \textbf{Zero-Hit Regime:} Under a rollout budget, sampled trajectories from $\pi_\text{LLM}$ miss the optimal basin. (b) \textbf{LatentRevise:} Starting from a failed rollout, we optimize its reasoning prefix in latent space via position-wise correction signals ($\nabla$), turning the original path into a successful one.} 
  \label{fig:intro}
\end{figure}

The boundary case is \emph{zero-hit reasoning under a rollout budget}, illustrated in Figure~\ref{fig:intro}(a): a prompt with pass@$N=0$, where none of the $N$ rollouts passes the verifier. This does not imply that the prompt is unsolvable; a verified trajectory may exist, but have too little probability mass under the current policy to be sampled within the budget. Such prompts are especially valuable because they lie at the finite-budget sampling frontier where new reasoning behavior is most needed. Yet standard RLVR often discards these zero-hit prompts rather than learning from them. With all rewards identical, group-normalized advantages collapse, so a zero-hit group provides no constructive update.

Existing remedies fall into two broad categories. One direction is to push the sampling frontier by increasing the rollout budget at higher cost, or by raising sampling randomness~\citep{hou_t1_2025,yu_dapo_2025}. However, these methods remain outcome-only: the verifier provides a binary score for the whole trajectory, without indicating which intermediate decision should change. As reasoning chains lengthen and correct solutions become rarer, searching becomes increasingly inefficient.
Another direction is to introduce dense guidance through on-policy distillation, where a teacher provides token-level targets on student-sampled trajectories~\citep{fan_on-policy_2025}; recent self-distillation variants further condition the teacher on privileged traces or rich feedback~\citep{zhao_self-distilled_2026,hubotter_reinforcement_2026}. These methods address the locality missing from outcome rewards, but depend on high-quality privileged information or feedback and reliable teacher guidance. When this guidance is weak or mismatched, it does not by itself turn a zero-hit group into useful supervision~\citep{hubotter_reinforcement_2026,fu_revisiting_2026,li_rethinking_2026}.

We introduce LatentRevise, a first-order latent revision method to learn from zero-hit prompts by revising their failed rollouts. LatentRevise treats failed rollouts not as dead ends but as on-policy traces that expose the model's reasoning structure and its intermediate errors. Rather than drawing more samples in the same discrete token space, it freezes the model and optimizes the failed reasoning prefix in latent space using dense, position-wise correction signals from the gold answer and its own failed continuation (Figure~\ref{fig:intro}(b)), without relying on teacher-provided token targets. To prevent optimization from drifting into adversarial feature directions, each update is constrained to the convex hull of the model's vocabulary embeddings through a Frank-Wolfe step, keeping the latent close to real token embeddings. 
The revised prefix is rolled out again, and the soft prefix is projected into discrete tokens; successful trajectories are retained as supervised data or injected into RL groups that would otherwise be ignored.

Empirically, LatentRevise rescues a notable fraction of zero-hit prompts by recovering at least one successful trajectory, turning them learnable where standard baselines obtain little benefit. This yields substantial improvements on math reasoning benchmarks, especially at larger pass@$k$, suggesting that recoverable zero-hit prompts are high-value training data rather than noise. The recovered continuations are also longer, more self-reflective, and reach correct answers missed by the original rollouts, indicating that latent revision surfaces error-correction behaviors the model can express but rarely reaches by discrete sampling alone.

Our contributions are threefold:
\begin{itemize}
    \item We frame zero-hit prompts, often discarded by RLVR pipelines, as a recoverable frontier: missed within the rollout budget, yet reachable from the model's own failed rollouts.

    \item We propose LatentRevise, a first-order latent revision method that turns failed rollouts into successful ones without external reference traces or a stronger teacher.

    \item We show that recovered trajectories alone are valuable training data, improving both pass@$1$ and pass@$k$ on math reasoning benchmarks.
\end{itemize}

\section{Related Work}
\label{sec:related}
\textbf{RLVR under Support Constraints.}
Self-training methods improve reasoning by learning from the model's own rollouts: STaR~\citep{zelikman_star_2022} and RFT~\citep{yuan_scaling_2023} train on sampled correct trajectories, while group-relative RLVR methods~\citep{shao_deepseekmath_2024,yu_dapo_2025} reweight current-policy rollouts by verifier scores. Recent analyses characterize this update as support-constrained: RLVR can amplify behavior already reached by sampling, but cannot reinforce verified solutions absent from the empirical group, and empirical support may even narrow during training~\citep{wu_invisible_2025,yue_does_2025}. Exploration-side remedies such as larger rollout budgets, higher decoding temperature, and entropy regularization~\citep{hou_t1_2025,yu_dapo_2025} can push the sampling frontier outward, but still operate through outcome-scored samples and leave a regime beyond reach under a practical budget. LatentRevise targets this regime by revising failed rollouts into verifier-passing trajectories that can feed SFT or RLVR.

\textbf{On-Policy Distillation under Privileged Conditioning.}
\citet{fan_on-policy_2025} observe that RLVR's trajectory-level correctness signal provides only coarse credit assignment on long reasoning chains, and propose on-policy distillation (OPD) as a dense replacement that supervises student-sampled rollouts with token-level teacher targets. Recent self-distillation variants obtain such teachers by conditioning the model on privileged information: OPSD conditions the teacher on reference reasoning traces~\citep{zhao_self-distilled_2026}, and SDPO conditions it on rich textual feedback or self-generated correct traces~\citep{hubotter_reinforcement_2026}. 
Recent analyses show that OPD can fail when teacher guidance becomes unreliable on student-generated states, when teacher and student reasoning patterns are mismatched, or when the teacher adds little genuinely new information~\citep{fu_revisiting_2026,li_rethinking_2026}.
LatentRevise instead derives position-wise guidance without a teacher distribution, using the gold answer and failed continuation to optimize the student's failed prefix in its own latent space.

\textbf{Latent Optimization for Reasoning.}
A parallel line uses continuous-space optimization to guide reasoning at inference time. LatentSeek~\citep{li_seek_2025} performs test-time instance-level policy-gradient updates on latent representations, guided by self-generated reward signals. SEA~\citep{wang_nabla-reasoner_2026} performs gradient-based sampling in continuous latent space against an energy function defined by the optimal policy, for inference-time alignment. DTO~\citep{yuan_inference-time_2025} applies differentiable optimization over token logits inside the decoding loop, using gradient signals from both a reward model and the language-model likelihood to iteratively refine the policy during decoding. These methods deploy continuous optimization as a per-instance inference-time procedure. LatentRevise uses latent optimization for a different purpose: recovering training signal when a rollout budget finds no positive trajectory, by optimizing a failed rollout in latent space. This training-time focus distinguishes LatentRevise from inference-time latent methods and connects it more directly to the exploration problem in RLVR.

\begin{figure*}
    \centering

    \includegraphics[width=1\linewidth]{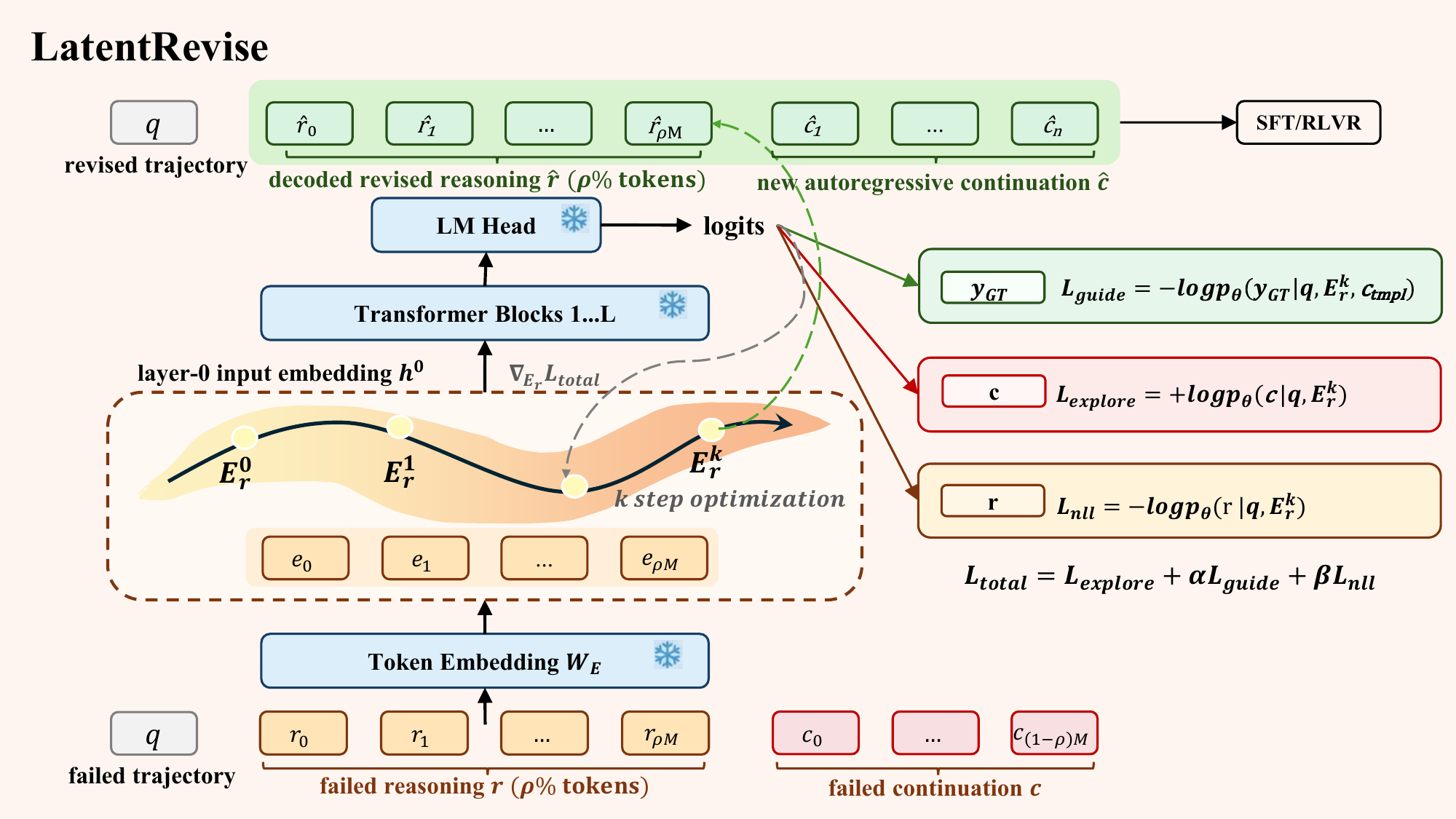}
    \caption{Pipeline of LatentRevise. Starting from a failed rollout $y_{\mathrm{fail}}=[r;c]$, LatentRevise optimizes the input embeddings of the reasoning prefix $r$ while keeping the model frozen. At step $k$, the latent prefix $E_r^{(k)}$ receives three gradient signals: guidance from the gold answer $y_{\mathrm{GT}}$, repulsion from the failed continuation $c$, and a fluency prior from the original prefix $r$. The optimized soft prefix decodes a candidate continuation, and only verified hard trajectories are retained for downstream SFT or RLVR.}
    \label{fig:framework}
\end{figure*}

\section{Methodology}
\label{sec:method}

\subsection{Preliminaries: Zero-Hit Regime in RL}
\label{subsec:preliminary}

We consider a reasoning prompt $q$ and a language-model policy $\pi_\theta$. In group-based RLVR methods such as GRPO~\citep{shao_deepseekmath_2024}, the policy samples a group of $N$ trajectories $\{y_1,\ldots,y_N\}$ and receives verifier rewards $R_i$. A common normalized advantage is 
\begin{equation}
    A_i = \frac{R_i - \bar{R}}{s_R + \varepsilon_{\mathrm{adv}}},
\end{equation}
where $\bar{R}$ and $s_R$ denote the group mean and standard deviation, and $\varepsilon_{\mathrm{adv}}>0$ is used for numerical stability. 

However, this mechanism fundamentally struggles in the hard exploration regime. A prompt is zero-hit under the current rollout budget when pass@$N=0$, i.e., none of the sampled trajectories passes the verifier. For binary verifier rewards, all failed trajectories receive the same reward, so $R_i=\bar{R}$ for every $i$ and the normalized advantage is zero, depriving the policy update of a meaningful gradient signal from these failed rollouts. This work targets this zero-hit regime, aiming to algorithmically recover valid training signals from these failed examples.

\subsection{Overview of LatentRevise}
\label{subsec:overview}
To break the exploration deadlock, we introduce LatentRevise, a first-order latent revision framework that performs continuous optimization in the input-embedding space to revise failed reasoning trajectories. As illustrated in Fig.~\ref{fig:framework}, the overall pipeline operates as follows:

Given a complex query $q$, the model first generates an initially failed response $y_{\mathrm{fail}}=[r;c]$, where $r$ is the reasoning prefix (the initial $\rho$ fraction of the rollout) and $c$ is the subsequent continuation. Rather than discarding this rollout, LatentRevise represents the reasoning prefix through its input embeddings $E_r$ and optimizes them while keeping the model frozen. At each step, the current latent prefix $E_r^{(k)}$ receives three teacher-forced signals: guidance from the gold answer $y_{\mathrm{GT}}$, repulsion from the original failed continuation $c$, and a fluency prior from the original prefix $r$ (Sec.~\ref{subsec:latent_revise}).

To preserve semantic integrity during this update, we introduce a vocabulary-level manifold constraint to prevent shortcut learning (Sec.~\ref{subsec:frank_wolfe}). Finally, the optimized soft prefix $E_r^*$ is used to decode a candidate continuation; after projection back to hard tokens and verification, the recovered hard trajectory is retained for downstream SFT or RLVR (Sec.~\ref{subsec:discrete_recovery}).

\subsection{Objective of Latent Optimization}
\label{subsec:latent_revise}

To effectively steer the reasoning-prefix latent $E_r$ while preserving its linguistic structure, we formulate a composite objective with three criteria. Here, $P_\theta(\cdot\mid q,E_r,\cdot)$ denotes the next-token distribution conditioned on the optimized reasoning-prefix embeddings.

\textbf{Latent Exploration Loss ($\mathcal{L}_{\text{explore}}$):} To encourage the trajectory to move away from its original failed continuation $c$, we apply a repulsive signal that lowers the likelihood of reproducing this continuation. We define $\mathcal{L}_{\text{explore}}$ as the log-likelihood of generating $c$ given the current reasoning-prefix latent $E_r$:

\begin{equation}
    \mathcal{L}_{\text{explore}} = \sum_{t=1}^{|c|} \log P_\theta \big( c^{(t)} \mid q, E_r, c^{(<t)} \big)
\end{equation}
Because this term is the negative of the usual cross-entropy on $c$, minimizing $\mathcal{L}_{\text{explore}}$ decreases the probability of the original failed continuation, encouraging the latent state to move toward alternative solution paths.

\textbf{Latent Guidance Loss ($\mathcal{L}_{\text{guide}}$):} To steer the trajectory toward the correct answer, we append a fixed answer template $c_{\text{tmpl}}$ (e.g., ``\textit{Based on my current reasoning process, the final answer is}'') after the reasoning prefix and compute the standard cross-entropy loss exclusively on the gold answer $y_{\text{GT}}$:

\begin{equation}
    \mathcal{L}_{\text{guide}} = - \sum_{t=1}^{|y_{\text{GT}}|} \log P_\theta \big( y_{\text{GT}}^{(t)} \mid q, E_r, c_{\text{tmpl}}, y_{\text{GT}}^{(<t)} \big)
\end{equation}

\textbf{Language Fluency Prior ($\mathcal{L}_{\text{NLL}}$):} To ensure that the optimized latent $E_r$ retains the structural coherence of natural language and does not degenerate into adversarial noise, we compute the negative log-likelihood (NLL) over the original reasoning tokens $r$:
\begin{equation}
    \mathcal{L}_{\text{NLL}} = - \sum_{t=1}^{|r|} \log P_\theta \big( r^{(t)} \mid q, E_r^{(<t)} \big)
\end{equation}

The total objective is the weighted sum of these three components:
\begin{equation}
    \mathcal{L}_{\text{total}} = \mathcal{L}_{\text{explore}} + \alpha \mathcal{L}_{\text{guide}} + \beta \mathcal{L}_{\text{NLL}}
\end{equation}
where $\alpha$ and $\beta$ are hyperparameters scaling the guidance and fluency terms.

\subsection{Manifold-Constrained Optimization}
\label{subsec:frank_wolfe}

Directly minimizing $\mathcal{L}_{\text{total}}$ with unconstrained optimizers (e.g., Adam) can drive the continuous representation $E_r$ away from the model's discrete token-embedding geometry, enabling the model to trivially ``cheat'' by encoding the answer into uninterpretable feature vectors. To mitigate this, we constrain the optimization to the convex hull of the model's vocabulary embeddings $W \in \mathbb{R}^{|V| \times d}$, utilizing the Frank-Wolfe (Conditional Gradient) algorithm. In complex mathematical reasoning, the pivotal token necessary to correct a trajectory (e.g., a specific operator) might fall outside the top-$K$ probable tokens; therefore, rather than restricting the search space, we intentionally search across the entire vocabulary $V$.

At each optimization iteration $i$, we compute the gradient $G = \nabla_{E_r} \mathcal{L}_{\text{total}} \in \mathbb{R}^{|r| \times d}$. For the embedding at position $j$, the Frank-Wolfe algorithm identifies the optimal vertex $v_j^*$ in the vocabulary matrix $W$ that minimizes the linear approximation of the loss and the latent embedding is then updated via a convex combination:
\begin{gather}
    v_j^* = \arg\min_{v \in W} \, \langle G_j, v \rangle \\
    E_r = (1 - \gamma_i) E_r + \gamma_i V^*
\end{gather}
We define a monotonically decaying step size $\gamma_i = \gamma_0 \frac{2}{i+1}$, which allows broad exploration in early steps and fine-grained convergence in later steps. We include a simple early-stopping criterion  to prevent over-optimization.  The complete optimization procedure is summarized in Algorithm~\ref{alg:fw_optimization}.

\begin{algorithm}[t!]
\caption{LatentRevise}
\label{alg:fw_optimization}
\begin{algorithmic}[1]
\REQUIRE Query $q$, failed rollout $y_{\text{fail}} = [r; c]$, gold answer $y_{\text{GT}}$, answer template $c_{\text{tmpl}}$, LLM $\theta$, vocabulary embedding matrix $W$, max steps $K$, stop threshold $\epsilon$.
\STATE Initialize reasoning latent $E_r \leftarrow \text{Embed}_\theta(r)$.
\FOR{$i = 1$ \TO $K$}
    \STATE $\mathcal{L}_{\text{explore}} \leftarrow \sum \log P_\theta(c \mid q, E_r)$
    \STATE $\mathcal{L}_{\text{guide}} \leftarrow - \sum \log P_\theta(y_{\text{GT}} \mid q, E_r, c_{\text{tmpl}})$
    \STATE $\mathcal{L}_{\text{NLL}} \leftarrow - \sum \log P_\theta(r \mid q, E_r)$

    \STATE $\mathcal{L}_{\text{total}} \leftarrow \mathcal{L}_{\text{explore}} + \alpha \mathcal{L}_{\text{guide}} + \beta \mathcal{L}_{\text{NLL}}$

    \STATE \textbf{if} $\mathcal{L}_{\text{guide}} < \epsilon$ \textbf{then break} \hfill \textit{\small $\triangleright$ Stop criterion}
    
    \STATE $\gamma_i \leftarrow \gamma_0 \left(\frac{2}{i+1}\right)$ 
    
    \STATE $V^* \leftarrow \arg\min_{V \in W^{|r|}} \text{Tr}\big( (\nabla_{E_r} \mathcal{L}_{\text{total}}) V^\top \big)$ 
    \STATE $E_r \leftarrow (1 - \gamma_i) E_r + \gamma_i V^*$
\ENDFOR
\RETURN Optimized reasoning latent $E_r$
\end{algorithmic}
\end{algorithm}

\subsection{Discrete Trajectory Recovery}
\label{subsec:discrete_recovery}

To utilize the optimized latent prefix $E_r^*$ in standard SFT or RL pipelines, we convert it into a discrete ``hard'' trajectory. First, we use $E_r^*$ as a soft prefix to autoregressively decode a new continuation $\hat{c}$. Because the reasoning prefix is supplied by $E_r^*$ rather than autoregressively regenerated, decoding only generates the continuation, introducing little computational overhead. We evaluate the decoded continuation against the gold answer $y_{\text{GT}}$ and discard any trajectory that fails to output the correct answer.

For successfully verified soft prefix rollouts, we project the optimized prefix back into discrete tokens. At each position $j$, we perform a nearest-neighbor search to identify the token ID that maximizes the inner product with the vocabulary matrix $W$:
\begin{equation}
    \hat{r}^{(j)} = \arg\max_{v_k \in W} \langle E_r^{*(j)}, v_k \rangle
\end{equation}
This operation discretizes the optimized latents into a hard reasoning sequence $\hat{r}$. The final concatenated trajectory $[\hat{r}; \hat{c}]$ provides a structurally coherent error-recovery path ready for SFT or inserted into RLVR groups.

\begin{table*}[t!]
\centering
\caption{Pass@k (\%) under DAPO-14k-Hard training}
\label{tab:hard_results}
\resizebox{\textwidth}{!}{%
\begin{tabular}{ll cc cc cc cc cc}
\toprule
\multirow{2}{*}{Model} & \multirow{2}{*}{Method} & \multicolumn{2}{c}{AIME 24} & \multicolumn{2}{c}{AIME 25} & \multicolumn{2}{c}{AMC 23} & \multicolumn{2}{c}{Math-500} & \multicolumn{2}{c}{Average} \\
\cmidrule(lr){3-4} \cmidrule(lr){5-6} \cmidrule(lr){7-8} \cmidrule(lr){9-10} \cmidrule(lr){11-12}
& & @1 & @32 & @1 & @32 & @1 & @32 & @1 & @32 & @1 & @32 \\
\midrule
\multirow{6}{*}{\shortstack{Qwen2.5\\1.5B-Inst}}
& Base        & 0.42 & 10.00 & 0.94 & 20.00 & 22.34 & 72.50 & 40.49 & 84.20 & 16.05          & 46.67          \\
& GRPO       & 0.42 & 6.67  & 0.62 & 16.67 & 22.89 & 75.00 & 34.66 & 81.00 & 14.65 & 44.84 \\
& GRPO(n=16) & 0.73 & 16.67 & 0.83 & 16.67 & 25.08 & 85.00 & 44.21 & 84.40 & 17.71 & 50.69 \\
& OPSD       & 0.31 & 10.00 & 0.31 & 10.00 & 9.84  & 67.50 & 17.81 & 74.20 & 7.07  & 40.42 \\
& SFT+Ours    & 0.73 & 13.33 & 0.78 & 26.67 & 25.00 & 75.00 & 43.97 & 87.00 & 17.62 & 50.50 \\
& GRPO+Ours   & 1.12& 16.67 & 1.04& 26.67 & 25.00 & 82.50 & 45.35 & 88.20& 18.13& 53.51 \\
\midrule
\multirow{6}{*}{\shortstack{Qwen3\\8B-Base}}
& Base        & 4.79 & 33.33 & 5.00  & 43.33 & 29.45 & 90.00 & 33.73 & 90.80 & 18.24          & 64.36          \\
& GRPO       & 5.21& 30.00& 5.94& 40.00& 31.02& 90.00& 35.56& 91.00& 19.43& 62.75\\
& GRPO(n=16) & 5.31& 30.00& 5.31& 30.00& 33.67& 85.00& 39.76& 90.60& 21.01& 58.90\\
& OPSD       & 2.92 & 30.00& 4.06  & 33.33 & 26.17 & 87.50 & 31.21 & 90.60 & 16.09 & 60.36 \\
& SFT+Ours    & 5.41& 33.33& 6.53& 43.33& 34.86& 90.00& 38.27& 91.20& 21.26& 64.46\\
& GRPO+Ours   & 5.94& 33.33& 7.12& 43.33& 36.67& 92.50& 40.59& 91.40& 22.58 & 65.14 \\
\bottomrule
\end{tabular}}
\end{table*}

\begin{table*}[t!]
\centering
\caption{Pass@k (\%) under DAPO-14k-Full training.}
\label{tab:full_results}
\resizebox{\textwidth}{!}{%
\begin{tabular}{ll cc cc cc cc cc}
\toprule
\multirow{2}{*}{Model} & \multirow{2}{*}{Method} & \multicolumn{2}{c}{AIME 24} & \multicolumn{2}{c}{AIME 25} & \multicolumn{2}{c}{AMC 23} & \multicolumn{2}{c}{Math-500} & \multicolumn{2}{c}{Average} \\
\cmidrule(lr){3-4} \cmidrule(lr){5-6} \cmidrule(lr){7-8} \cmidrule(lr){9-10} \cmidrule(lr){11-12}
& & @1 & @32 & @1 & @32 & @1 & @32 & @1 & @32 & @1 & @32 \\
\midrule
\multirow{5}{*}{\shortstack{Qwen2.5\\1.5B-Inst}}
& Base        & 0.42 & 10.00 & 0.94 & 20.00 & 22.34 & 72.50 & 40.49 & 84.20 & 16.05          & 46.67          \\
& GRPO        & 0.83& 10.00& 0.62& 13.33& 23.98& 77.50& 46.22& 85.40& 17.91 & 46.56 \\
& GRPO(n=16)& 0.83& 16.67& 0.42& 10.00& 23.20& 72.50& 46.87& 85.40& 17.83 & 46.14 \\
& DAPO        & 0.31& 10.00& 0.83& 10.00& 24.53& 77.50& 46.67& 86.80& 18.09 & 46.08 \\
& Ours& 0.83& 20.00& 1.04& 20.00& 25.39& 82.50& 47.54& 87.50& 18.70 & 52.50 \\
\midrule
\multirow{5}{*}{\shortstack{Qwen3\\8B-Base}}
& Base        & 4.79 & 33.33 & 5.00  & 43.33 & 29.45 & 90.00 & 33.73 & 90.80 & 18.24          & 64.36          \\
& GRPO        & 8.54& 33.33& 11.15& 33.33& 49.30& 97.50& 62.04& 92.00& 32.76& 64.04\\
& GRPO(n=16)& 7.50& 43.33& 12.19& 46.67& 52.03& 92.50& 65.59& 92.80& 34.33& 68.83\\
& DAPO        & 7.81& 40.00& 10.94& 36.67& 46.09& 95.00& 59.23& 92.60& 31.02& 66.07\\
& Ours& 9.39& 43.33& 12.54& 50.00& 53.19& 97.50& 67.94& 94.20& 35.77 & 71.26 \\
\bottomrule
\end{tabular}}
\end{table*}
\section{Experiments}
\label{sec:experiments}

\subsection{Experimental Setup}
\label{subsec:exp_setup}

\textbf{Models, Datasets, and Baselines.} We evaluate Qwen2.5-1.5B-Instruct~\citep{qwen_qwen25_2024} and Qwen3-8B-Base~\citep{yang_qwen3_2025} on DAPO-Math-14K~\citep{yu_dapo_2025} under two settings: \textit{dapo-hard} (queries where the base model achieves pass@32\,=\,0) and \textit{dapo-full} for end-to-end RL. Evaluation covers MATH500~\citep{hendrycks_measuring_2021}, AIME 24, AIME 25, and AMC 23, OlympiadBench\citep{he2024olympiadbench} and Minerva Math\citep{lewkowycz2022solving}. Baselines include GRPO~\citep{shao_deepseekmath_2024}, DAPO~\citep{yu_dapo_2025}, OPSD~\citep{zhao_self-distilled_2026} with the gold answer as privileged information, and GRPO ($N\!=\!16$) as a stronger sampling baseline. Unless otherwise noted, we use rollout group size $N=8$ and run LatentRevise for $K=10$ Frank-Wolfe steps over the first $\rho=0.8$ fraction of the reasoning prefix. Full details are in Appendix~\ref{sec:appendix_setup}.Addition Results on OlympiadBench and Minerva Math are in Appendix~\ref{sec:appendix_extra_results}

\textbf{LatentRevise Integration.} In the \textbf{offline SFT} setting, we apply latent revision to failed trajectories in \textit{dapo-hard}, retain recovered correct paths, and fine-tune via cross-entropy. In the \textbf{on-policy RL} setting, LatentRevise acts as a dynamic fallback: whenever standard sampling yields no correct rollout for a query, we revise a failed path and inject the recovered trajectory back into the group before recomputing advantages. This design requires no changes to the RL objective and introduces no additional cost for queries that already have at least one correct rollout.

\begin{table*}[t]
  \caption{Ablation study and sensitivity analysis on LatentRevise (Qwen2.5-1.5B-Instruct, DAPO-Hard). Results show Pass@$k$ (\%).}
  \label{tab:ablation_sensitivity}
  \centering
  \resizebox{\textwidth}{!}{%
  \begin{tabular}{ll cc cc cc cc cc}
    \toprule
    \multirow{2}{*}{\textbf{Config}} & \multirow{2}{*}{\textbf{}} & \multicolumn{2}{c}{AIME 24} & \multicolumn{2}{c}{AIME 25} & \multicolumn{2}{c}{AMC 23} & \multicolumn{2}{c}{Math-500} & \multicolumn{2}{c}{Average} \\
    \cmidrule(lr){3-4} \cmidrule(lr){5-6} \cmidrule(lr){7-8} \cmidrule(lr){9-10} \cmidrule(lr){11-12}
    & & @1 & @32 & @1 & @32 & @1 & @32 & @1 & @32 & @1 & @32 \\
    \midrule
    \multicolumn{12}{l}{\textit{Ablation Study}} \\
    \textbf{Full LatentRevise}              & & 1.12 & 16.67 & 1.04 & 26.67 & 25.00 & 82.50 & 45.35 & 88.20 & 18.13 & 53.51 \\
    \quad w/o FW Optimizer                  & & 0.21 & 3.33  & 0.21 & 3.33  & 4.69  & 42.50 & 10.16 & 55.40 & 3.82  & 26.14 \\
    \quad w/o $\mathcal{L}_{\text{guide}}$  & & 0.52 & 10.00 & 0.52 & 13.33 & 14.45 & 62.50 & 28.67 & 72.40 & 11.04 & 39.56 \\
    \quad w/o $\mathcal{L}_{\text{explore}}$& & 1.04 & 13.33 & 0.94 & 23.33 & 23.44 & 80.00 & 43.75 & 85.40 & 17.29 & 50.52 \\
    \quad w/o $\mathcal{L}_{\text{NLL}}$    & & 0.94 & 13.33 & 0.83 & 23.33 & 22.66 & 77.50 & 42.18 & 84.80 & 16.65 & 49.74 \\
    \midrule
    \multicolumn{12}{l}{\textit{Sensitivity: Thinking Ratio (Steps = 10)}} \\
    Thinking Ratio = 0.5 & & 0.94 & 13.33 & 0.83 & 20.00 & 22.27 & 77.50 & 41.93 & 84.60 & 16.49 & 48.86 \\
    Thinking Ratio = 0.2 & & 0.52 & 10.00 & 0.52 & 16.67 & 16.02 & 67.50 & 33.20 & 77.80 & 12.57 & 42.99 \\
    \midrule
    \multicolumn{12}{l}{\textit{Sensitivity: Opt.\ Steps (Ratio = 0.8)}} \\
    Opt.\ Steps = 5   & & 0.73 & 10.00 & 0.73 & 20.00 & 20.08 & 72.50 & 38.67 & 82.20 & 15.05 & 46.17 \\
    Opt.\ Steps = 20  & & 1.04 & 16.67 & 0.94 & 26.67 & 24.22 & 82.50 & 44.92 & 86.80 & 17.78 & 53.16 \\
    \bottomrule
  \end{tabular}}
\end{table*}
\begin{table}[t]
  \caption{Cost-performance analysis on Qwen2.5-1.5B-Inst. Time and memory measured on 4×B200 GPUs.}
  \label{tab:cost_perf}
  \centering
  \resizebox{\columnwidth}{!}{%
  \begin{tabular}{l ccc}
    \toprule
    \textbf{Metric} & GRPO & GRPO(N=16) & \textbf{Ours} \\
    \midrule
    Time (hr)  & 5.8 & 9.6 & 7.2 \\
    Mem.\ (GB) & 46.2 & 52.8 & 47.8 \\
    Avg.\ Pass@32 (\%) & 44.84 & 50.69 & 53.51\\
    \bottomrule
  \end{tabular}}
\end{table}

\subsection{Results on Zero-Hit Queries (DAPO-Hard)}
\label{subsec:zero_hit}

Table~\ref{tab:hard_results} compares LatentRevise against baselines on queries where the base model achieves pass@32\,=\,0.
The resulting subset contains 6,580 zero-hit questions for Qwen2.5-1.5B-Instruct and 3,642 for Qwen3-8B-Base, corresponding to 46.6\% and 25.8\% of the training set, respectively.
GRPO suffers from collapsed advantages on these queries and provides little benefit for improving the policy. OPSD bypasses the issue by conditioning the teacher on the ground-truth answer, but it suffers from hindsight bias and does not reliably produce useful token-level guidance for the student's failed reasoning states: its performance falls below the base model.
SFT+Ours provides supporting evidence that recovered trajectories are useful supervision, while \textbf{GRPO+Ours} further amplifies the benefit and achieves the highest Average pass@1 and pass@32 across both models.
This shows that revising recoverable frontier failures provides more useful training signal than either hindsight teacher conditioning or simply enlarging the rollout group.

A key finding from Table~\ref{tab:hard_results} is that training only on these zero-hit problems under a rollout budget yields substantial pass@$k$ gains. These gains indicate that recoverable zero-hit prompts are not merely outliers to discard: once revised into verified trajectories, they provide useful training signal for improving broader solution coverage. This suggests that hard, unsolved queries near the sampling frontier can be especially valuable training data for expanding model capability.

\subsection{Results on Full Datasets (DAPO-Full)}
\label{subsec:full}
Table~\ref{tab:full_results} evaluates LatentRevise in the full DAPO-Math-14K training setting, where zero-hit queries are mixed with easier and partially solved queries. The gains persist in this end-to-end setting: \textbf{Ours} outperforms GRPO, DAPO, and the larger-rollout GRPO baseline on both models, with the clearest improvements at pass@32. This shows that LatentRevise is not merely effective on a filtered hard subset. By recovering verified trajectories from zero-hit groups inside the full training distribution, it supplies complementary training signal that helps the model reach solutions missed by standard RL and additional sampling.


\subsection{Ablation Studies and Sensitivity Analysis}
\label{subsec:ablation}
\textbf{Ablation.} Table~\ref{tab:ablation_sensitivity} evaluates four ablated variants. Removing the Frank-Wolfe (FW) optimizer causes the largest degradation (pass@32: 53.51\%\,$\to$\,26.14\%), highlighting the importance of constraining each update toward vocabulary-embedding directions rather than allowing unconstrained latent drift. Removing $\mathcal{L}_{\text{guide}}$ yields the second-largest drop (39.56\%), showing that the gold-answer anchor provides the main directional signal for revision. Removing $\mathcal{L}_{\text{explore}}$ or $\mathcal{L}_{\text{NLL}}$ incurs modest degradation, indicating these terms provide meaningful regularization.

\begin{figure*}    \centering
    \includegraphics[width=1.0\linewidth]{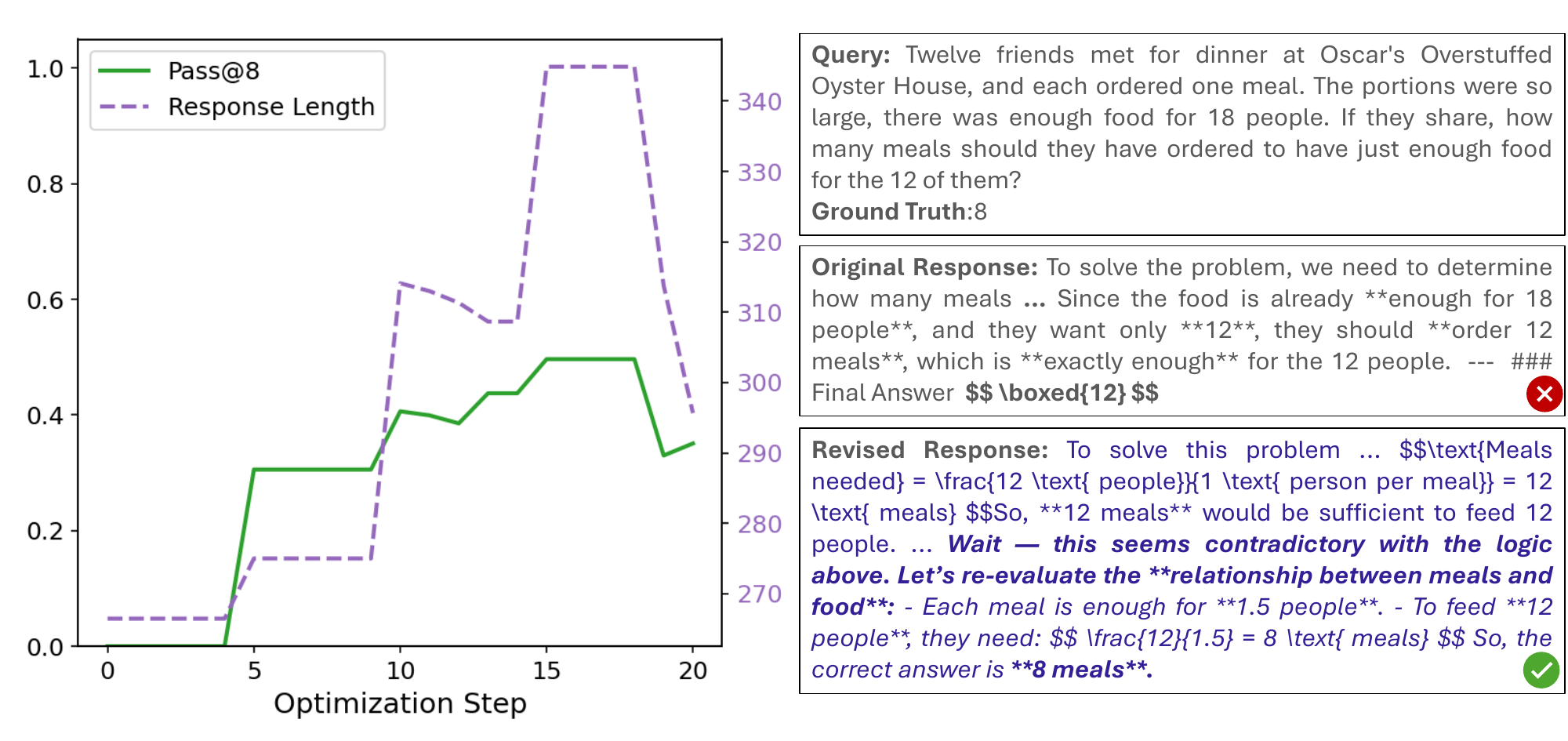}
    \caption{Optimization dynamics and a representative case study. \textbf{Left}: pass@8 rises sharply in the first 10 steps, with about 40\% of zero-hit prompts yielding at least one verified trajectory, before plateauing with mild fluctuations; generated continuation length also grows, suggesting additional reasoning steps. \textbf{Right}: the revised response revisits the key condition, corrects the interpretation, and reaches the verified answer.}
    \label{fig:opt_dynamics}
\end{figure*}
\textbf{Sensitivity.} Decreasing the thinking ratio from 0.8 to 0.5 or 0.2 monotonically reduces performance (48.86\% and 42.99\%), as fewer editable embeddings limit the optimizer's steering capacity. For optimization steps, 5 steps is insufficient, while 20 steps is comparable to the default 10-step setting, indicating rapid convergence.

\subsection{Analysis: Latent Optimization Dynamics}
\label{subsec:analysis}
Figure~\ref{fig:opt_dynamics} plots pass@8 and generated continuation length across optimization steps on Qwen2.5-1.5B-Instruct. Two trends emerge: \textbf{accuracy rises sharply in the first 10 steps} and then plateaus with mild fluctuations; and \textbf{response length grows beyond the original generation length}, reflecting the insertion of additional reasoning steps.

Closer inspection shows that revised trajectories frequently contain explicit self-reflection markers, such as ``wait, let me reconsider'' or ``actually, this step is incorrect'', which are absent from the original failed rollout. As illustrated in Figure~\ref{fig:opt_dynamics}, the latent-revised trajectory revisits the key condition after its initial incorrect answer and reaches the verified solution, without apparent fluency degradation in the decoded text. 
This behavior suggests that latent revision can expose recoverable error-correction patterns rather than only patching the final answer.



\subsection{Training Efficiency}
\label{subsec:efficiency}
Table~\ref{tab:cost_perf} shows the cost-performance tradeoff on Qwen2.5-1.5B-Instruct. Doubling the rollout group from $N=8$ to $N=16$ increases training time by about 66\% and memory by about 14\%, while LatentRevise adds a smaller overhead over GRPO, about 24\% more time and 3\% more memory. The lower overhead comes from the structure of LatentRevise: each revision step uses forward-backward passes, which are parallelizable matrix multiplications rather than autoregressive decoding, keeping cost proportional to the number of revision steps rather than generated sequence length. LatentRevise is both less costly than GRPO($N\!=\!16$) and achieves the best average pass@32, indicating that revising failed rollouts is more efficient than simply increasing the sampling budget. This efficiency advantage becomes more pronounced as model scale increases, since autoregressive generation cost grows with both sequence length and model size.

\section{Conclusion}
\label{sec:conclusion}

When a model fails to reach a correct trajectory under a practical rollout budget, the usual remedies are to sample more or introduce stronger external guidance, such as a teacher, reference trace, or process supervision. We ask whether the same model can instead recover training signal from its own failed rollouts. LatentRevise, our first-order latent revision framework, gives a positive answer for recoverable zero-hit prompts: by optimizing the input embeddings of a failed reasoning prefix under vocabulary-constrained updates, it turns frontier failures into verified trajectories usable as ordinary training data. Across both filtered hard subsets and full-dataset training, LatentRevise improves over standard RL and larger-rollout controls, with the clearest gains at pass@32 when integrated into on-policy RL.


More broadly, LatentRevise suggests a complementary axis for reasoning RL: when token-level sampling reaches its finite-budget frontier, first-order revision in the model's own latent space can still turn failed rollouts into verified supervision. As reasoning tasks become harder and correct trajectories require longer chains, this axis turns a growing boundary of sampling-based training into a source of curriculum and training signal.



\bibliography{my_custom}

\clearpage
\appendix

\section{Detailed Experimental Setup}
\label{sec:appendix_setup}

\paragraph{Datasets.}
We train on DAPO-Math-14K~\citep{yu_dapo_2025}, the English version of DAPO-Math-17K, which contains approximately 14{,}000 competition-style mathematical problems spanning algebra, number theory, combinatorics, and geometry. For the \textit{DAPO-Hard} setting, we filter this dataset to retain only queries where the base model achieves pass@32\,=\,0 under standard autoregressive sampling. Evaluation is performed on four benchmarks: AIME 2024~\citep{aime24}, AIME 2025~\citep{aime25}, AMC 2023, MATH500~\citep{hendrycks_measuring_2021}, OlympiadBench and Minerva Math. All datasets used in this work are publicly available and used in accordance with their respective licenses for research purposes.

\section{Results on Additional Mathematical Benchmarks}
\label{sec:appendix_extra_results}

\begin{table}[t]
\centering
\caption{Results on OlympiadBench and Minerva Math (Hard subset training).}
\label{tab:extra_hard}
\resizebox{0.5\textwidth}{!}{
\begin{tabular}{lcccc}
\toprule
\multirow{2}{*}{Method} & \multicolumn{2}{c}{OlympiadBench} & \multicolumn{2}{c}{Minerva Math} \\
\cmidrule(lr){2-3} \cmidrule(lr){4-5}
 & pass@1 & pass@32 & pass@1 & pass@32 \\
\midrule
\multicolumn{5}{l}{\textit{Qwen2.5-1.5B-Instruct}} \\
Base & 11.2 & 37.4 & 6.4 & 27.6 \\
GRPO (n=8) & 11.5 & 34.9 & 7.1 & 29.0 \\
GRPO (n=16) & 11.7 & 36.2 & 7.8 & 32.0 \\
OPSD & 0.2 & 4.7 & 0.4 & 6.2 \\
SFT + Ours & 11.5& 35.3& 7.7& 31.7\\
GRPO + Ours & 11.9& 38.2& 8.1& 33.1\\
\midrule
\multicolumn{5}{l}{\textit{Qwen3-8B-Base}} \\
Base & 14.2& 45.1& 6.2& 36.7\\
GRPO (n=8) & 15.8 & 45.7 & 6.5 & 35.3 \\
GRPO (n=16) & 17.4 & 46.0 & 7.4 & 40.8 \\
OPSD & 12.9 & 43.0 & 6.2 & 35.3 \\
SFT + Ours & 16.9& 45.3& 6.7& 36.6\\
GRPO + Ours & 18.0& 47.2& 7.9& 41.2\\
\bottomrule
\end{tabular}
}
\end{table}

\begin{table}[t]
\centering
\caption{Results on OlympiadBench and Minerva Math (Full subset training).}
\label{tab:extra_full}
\resizebox{0.5\textwidth}{!}{
\begin{tabular}{lcccc}
\toprule
\multirow{2}{*}{Method} & \multicolumn{2}{c}{OlympiadBench} & \multicolumn{2}{c}{Minerva Math} \\
\cmidrule(lr){2-3} \cmidrule(lr){4-5}
 & pass@1 & pass@32 & pass@1 & pass@32 \\
\midrule
\multicolumn{5}{l}{\textit{Qwen2.5-1.5B-Instruct}} \\
Base & 11.2 & 37.4 & 6.4 & 27.6 \\
GRPO (n=8) & 11.8 & 35.6 & 8.6 & 30.5 \\
GRPO (n=16) & 12.0 & 36.8 & 8.7 & 30.5 \\
DAPO & 11.9 & 36.1 & 8.7 & 29.4 \\
Ours& 12.4& 38.0& 9.1& 31.9\\
\midrule
\multicolumn{5}{l}{\textit{Qwen3-8B-Base}} \\
Base & 14.2& 45.1& 6.2& 36.7\\
GRPO (n=8) & 24.6 & 47.5 & 12.5 & 41.9 \\
GRPO (n=16) & 25.7 & 46.7 & 12.7 & 39.3 \\
DAPO & 23.7 & 46.3 & 11.4 & 41.2 \\
Ours& 26.6& 48.9& 13.0& 42.1\\
\bottomrule
\end{tabular}
}
\end{table}

Tables \ref{tab:extra_hard} and \ref{tab:extra_full} extend our evaluation to OlympiadBench and Minerva Math. Under the zero-hit subset training (Table \ref{tab:extra_hard}), OPSD struggles to provide reliable guidance and degrades performance on Qwen2.5-1.5B-Instruct. In contrast, our method (\textbf{GRPO + Ours}) consistently achieves the highest pass@1 and pass@32 across both models, outperforming the larger-rollout GRPO ($n=16$) baseline. The strong performance of \textbf{SFT + Ours} further validates the high quality of our recovered trajectories. These gains persist in the full training setting (Table \ref{tab:extra_full}), where \textbf{Ours} consistently surpasses standard GRPO, DAPO, and GRPO ($n=16$). Notably, on Qwen3-8B-Base, our approach reaches peak pass@1 scores of 26.6 on OlympiadBench and 13.0 on Minerva Math. These results confirm that recovering verified trajectories from zero-hit queries provides robust, complementary training signals that effectively enhance broader mathematical reasoning capabilities.

\paragraph{Models.}
We experiment with two language models of different scales and families.
\begin{itemize}
  \item \textbf{Qwen2.5-1.5B-Instruct}~\citep{qwen_qwen25_2024}: a compact instruction-tuned model from the Qwen2.5 series, used to evaluate LatentRevise under a tight parameter budget.
  \item \textbf{Qwen3-8B-Base}~\citep{yang_qwen3_2025}: a larger base model from the Qwen3 family, used to assess whether the benefits of latent revision transfer to a stronger model with broader knowledge coverage.
\end{itemize}

\paragraph{Reward Function.}
We use a binary verifier reward: $+1$ if the model's final boxed answer matches the ground-truth answer, and $0$ otherwise. We intentionally exclude format-based auxiliary rewards, as they can constrain exploration and introduce reward hacking, particularly when training base models~\citep{yu_dapo_2025}.

\paragraph{Implementation Details.}
All models are trained using the \textsc{verl} framework~\citep{sheng_hybridflow_2024} with the GRPO algorithm. We use a training batch size of 512, a rollout group size of $N=8$, and a constant learning rate of $1\times10^{-5}$. The KL penalty coefficient is $\beta=0.001$ and the gradient clipping norm is 1.0. The maximum response length is set to 8{,}192 tokens during training and expanded to 32{,}000 during evaluation to accommodate long chain-of-thought reasoning. All experiments are conducted on nodes equipped with $4\times$ NVIDIA B200 GPUs.

For the LatentRevise optimization procedure, we run $K=10$ Frank-Wolfe steps per failed rollout, with a decayed step size starting at $\gamma_0=0.1$. The thinking ratio is $\rho=0.8$, meaning the first 80\% of the rollout (by token count) is treated as the reasoning prefix and subject to latent revision. Loss weights are $\alpha_{\text{guide}}=1.0$, and $\beta_{\text{NLL}}=0.1$. Only rollouts for which the revised soft prefix produces a verified correct continuation are retained; the remaining unrecoverable zero-hit examples are discarded and do not contribute to training.

\section{Hyperparameter Summary}
\label{sec:appendix_hyperparams}

Table~\ref{tab:hyperparams} consolidates all hyperparameters for reproducibility.

\begin{table}[h]
\centering
\caption{Hyperparameters for LatentRevise and downstream training.}
\label{tab:hyperparams}
\resizebox{\columnwidth}{!}{%
\begin{tabular}{lll}
\toprule
\textbf{Category} & \textbf{Hyperparameter} & \textbf{Value} \\
\midrule
\multirow{7}{*}{\shortstack{Latent\\Optimization}}
  & Optimization steps $K$                              & 10 \\
  & Thinking ratio $\rho$                               & 0.8 \\
  & Initial step size $\gamma_0$                        & 0.1 \\
  & Guide loss weight $\alpha$          & 1.0 \\
  & NLL regularizer weight $\beta$       & 0.1 \\
  & stop threshold $\epsilon$       & 1e-3 \\
\midrule
\multirow{9}{*}{\shortstack{GRPO\\Training}}
  & Learning rate                                       & $1\times10^{-5}$ \\
  & Training batch size                                 & 512 \\
  & Rollout group size $N$                              & 8 \\
  & Max sequence length (train)                         & 8192 \\
  & Max sequence length (eval)                          & 32000 \\
  & KL coefficient $\beta$                              & 0.001 \\
  & Gradient clip norm                                  & 1.0 \\
  & Optimizer                                           & AdamW \\
\bottomrule
\end{tabular}}
\end{table}

\section{Ethical Considerations}
\label{sec:appendix_ethics}

\noindent\textbf{Datasets and Licensing.} All datasets used in this work---DAPO-Math-14K, AIME 2024/2025, AMC 2023, MATH500, OlympiadBench and Minerva Math---are publicly released and used solely for non-commercial academic research, consistent with their intended purpose and licensing terms. No personally identifiable information is present in any of these datasets.

\noindent\textbf{Computational Resources.} Training large language models requires substantial computational resources. Our experiments were conducted on $4\times$ NVIDIA B200 GPUs. We encourage the community to continue developing compute-efficient alternatives to reduce the environmental footprint of large-scale RL training.

\noindent\textbf{Reliability of Model Outputs.} Mathematical reasoning models, including those trained with LatentRevise, may produce incorrect answers with apparent confidence. We do not advocate for deploying such models in high-stakes settings (e.g., automated grading or formal verification) without human oversight and rigorous evaluation.

\noindent\textbf{Dual-Use Considerations.} Improving the general reasoning capabilities of language models may enable beneficial applications in education and scientific discovery, but could also amplify the capacity of models to construct sophisticated but misleading arguments. We believe these risks are not specific to our method and are shared by the broader field of reasoning-focused post-training.

\section{AI Writing Assistance}
\label{sec:appendix_ai}

\noindent\textbf{Use of AI Assistance.} The writing of this paper was assisted by AI tools, primarily Claude (Anthropic). AI assistance was used for drafting and refining prose. All technical content, experimental design, results, and conclusions are the work of the authors.

\end{document}